\documentclass[sn-mathphys]{sn-jnl}

\usepackage{siunitx}
\usepackage[utf8]{inputenc}
\usepackage{graphicx,url}
\usepackage{amsmath}

\jyear{2021}%

\theoremstyle{thmstyleone}%
\theoremstyle{thmstyletwo}%

\theoremstyle{thmstylethree}%

\begin{document}

\title[Data vs classifiers, who wins?]{Data vs classifiers, who wins?}


\author*[1,3]{\fnm{Lucas} \sur{Cardoso}}\email{lucas.cardoso@icen.ufpa.br}

\author[1,3]{\fnm{Vitor} \sur{Santos}}\email{vitor.cirilo3@gmail.com}
\equalcont{These authors contributed equally to this work.}

\author[1]{\fnm{Regiane} \sur{Francês}}\email{kawasaki@ufpa.br}
\equalcont{These authors contributed equally to this work.}

\author[2]{\fnm{Ricardo} \sur{Prudêncio}}\email{rbcp@cin.ufpe.br}
\equalcont{These authors contributed equally to this work.}

\author*[3]{\fnm{Ronnie} \sur{Alves}}\email{ronnie.alves@itv.org}

\affil*[1]{\orgdiv{Faculdade de Computação}, \orgname{Universidade Federal do Pará}, \orgaddress{\street{Augusto Corrêa}, \city{Belém}, \postcode{66075-110}, \state{Pará}, \country{Brazil}}}

\affil[2]{\orgdiv{Centro de Informática}, \orgname{Universidade Federal de Pernambuco}, \orgaddress{\street{Ave Prof. Moraes Rego}, \city{Recife}, \postcode{50670-901}, \state{Pernambuco}, \country{Brazil}}}

\affil[3]{\orgdiv{Instituto Tecnológico Vale}, \orgname{Vale}, \orgaddress{\street{Boaventura da Silva}, \city{Belém}, \postcode{66055-090}, \state{Pará}, \country{Brazil}}}


\abstract{
The experiments covered by Machine Learning (ML) must consider two important aspects to assess the performance of a model: datasets and algorithms. Robust benchmarks are needed to evaluate the best classifiers. For this, one can adopt gold standard benchmarks available in public repositories. However, it is common not to consider the complexity of the dataset when evaluating. This work proposes a new assessment methodology based on the combination of Item Response Theory (IRT) and Glicko-2, a rating system mechanism generally adopted to assess the strength of players (e.g., chess). For each dataset in a benchmark, the IRT is used to estimate the ability of classifiers, where good classifiers have good predictions for the most difficult test instances. Tournaments are then run for each pair of classifiers so that Glicko-2 updates performance information such as rating value, rating deviation and volatility for each classifier. A case study was conducted hereby which adopted the OpenML-CC18 benchmark as the collection of datasets and pool of various classification algorithms for evaluation. Not all datasets were observed to be really useful for evaluating algorithms, where only 10\% were considered really difficult. Furthermore, the existence of a subset containing only 50\% of the original amount of OpenML-CC18 was verified, which is equally useful for algorithm evaluation. Regarding the algorithms, the methodology proposed herein identified the Random Forest as the algorithm with the best innate ability.
}




\maketitle

\section{Introduction}\label{sec1}

Machine Learning (ML) has been growing very fast in recent years, due to the large number of applications adopting ML models to support relevant tasks in different domains. The types of learning algorithms range from unsupervised to supervised \citep{monard2003conceitos}. This paper focuses on supervised learning algorithms, more precisely on classification algorithms, which are commonly adopted for pattern recognition tasks in several applications \citep{domingos2012few}. In literature, there is a wide range of supervised algorithms , which adopt different learning strategies to induce models from data \citep{monard2003conceitos}. Additionally, datasets can have particular characteristics, in such a way that it is not always simple to choose the best algorithm to be adopted for a given dataset. In this context, it is usual to empirically evaluate algorithms in order to identify the most promising ones for the datasets at hand. In fact, empirically evaluating ML algorithms is crucial to understand the advantages and limitations of the available techniques. 

Two important issues can be pointed out for empirically evaluating algorithms: (1) the datasets adopted in experiments; and (2) the very methodology of the experiments. Concerning the former issue, a common research practice in ML is to evaluate algorithms using benchmark datasets from online repositories, like the UCI respository \cite{Dua:2019}. Following the growth in the area, in recent years different online dataset repositories have emerged, such as OpenML \citep{vanschoren2014openml}. Specifically in OpenML, researchers can share datasets and experimental results, such as the performance of a classifier against a dataset. In addition, OpenML has several reference benchmarks, such as OpenML Curated Classification 2018 (OpenML-CC18), a benchmark for classification tasks that has 72 datasets curated and standardized by the platform \citep{bischl2017openml}.

Regarding the methodology of experiments, ML models can be trained and tested by applying a specific  execution procedure (e.g., cross-validation) on each dataset and then evaluated by using evaluation metrics of interest (e.g., accuracy). This strategy, however, does not allow for an in-depth analysis of the real ability of the model. Some datasets may be favoring an algorithm, giving the false impression that the classifier is, in fact, the best in relation to the others \citep{martinez2016making}. The complexity of the dataset should be taken into account during the process of evaluating the performance of an algorithm. According Kubat (2017) \citep{kubat2017introduction}, only the use of the classic evaluation metrics can not guarantee that the evaluation result is completely reliable. Therefore, it is important that other metrics are applied to result in a more robust assessment.

But in the data vs classifiers fight, which is more important? According to \cite{domingos2012few}, even a simpler classifier can beat the best classifier if the first classifier has much more training data than the second. At the same time, ``data alone is not enough'', i.e., the learning algorithms adopted in a dataset are crucial and make all the difference in the final learning results. Thus, data and models are two sides of the same coin of ML experimentation. So, an important question to address is how to evaluate classifiers, considering the dataset's importance as well. In previous works \citep{prudencio2015analysis,martinez2016making,martinez2019item,song2021efficient}, Item Response Theory (IRT) was adopted to answer this question and to provide a robust approach that allows for evaluating both datasets and classifiers. IRT has been widely adopted in psychometric tests to measure an individual's ability to correctly answer a set of items (e.g., questions) by
taking into account the difficulty of the items. High ability values are assigned for individuals that correctly solve the most difficult items, while maintaining a consistent performance in the easy items as well. By considering classifiers as individuals and test instances as items, it is then possible to apply the concepts of IRT in the ML field. Thus, when applying IRT in ML, it is possible to simultaneously evaluate datasets and algorithms already considering the complexity of the dataset when measuring the classifier performance.

Despite the advantages of applying IRT in ML, evaluating benchmarks and algorithms with IRT is still a non-straightforward task. In the current work, we propose the combination of IRT with rating systems \citep{samothrakis2014predicting}, in order to summarize the IRT results across datasets in a benchmark. Rating systems are commonly used to assess the ``strength'' of an individual in a competition (e.g., chess) through a series of tournaments. Rating systems are adopted to measure how proficient an individual is in a given activity. In this sense, such systems and IRT have some concepts in common which are explored herein. More specifically, this research adopted the Glicko-2 \citep{glickman2012example} rating system in order to create a ranking of models to summarize the results obtained by IRT on multiple datasets. According to the proposal of this research, each dataset is seen as a tournament in which the classifiers compete against each other. Initially, for each dataset, IRT is appled to measure the ability of each classifier. Then, each pair of classifiers compete against each other and a winner score is assigned to the classifier with higher ability for that dataset. The recorded scores are used by the Glicko-2 system to update the rating value, the rating deviation and the volatility of each classifier. After all datasets in the benchmark are considered in the tournaments, the final rating values are returned to produce a raking of classifiers.

In order to verify the viability of the methodology proposed hereby, a case study was conducted using the OpenML-CC18 benchmark. Preliminary results were obtained by \cite{cardoso2020decoding}, where a set of 60 datasets from the OpenML-CC18 benchmark was considered. In this research work, the case study was extended by addressing some questions like: Would it be possible to use the IRT estimators to choose the best benchmark subset within OpenML-CC18? Are there datasets within a benchmark that might not be really good for evaluating learning algorithms? For this, IRT is used to create subsets of OpenML-CC18 datasets, then the combination of IRT and Glicko-2 is applied to generate the classifier rankings. Next, each ranking is analyzed, considering the expected performance of the classifiers to evaluate and then choosing a possible subset that is more efficient than the original one. In addition, this research also seeks to use the rankings generated for each subset to explore, with the help of Glicko-2 system, the existence of an ``innate ability'' of the evaluated classifiers, and to provide guidance for selecting the winner algorithm.


The main contributions of this work are summarized as follows:

\begin{itemize}
    \item To propose a new methodology to simultaneously evaluate the performance of algorithms and the difficulty of datasets, based on the combination of IRT and Glicko-2;
     \item Application of the proposed methodology to analyze existing problems in a known benchmark in OpenML;
       \item To use the methodology proposed in OpenML-CC18 to point out the best classifier, thus exploring the concept of innate ability;
    \item All implementations in the case study are provided for in a single tool, called decodIRT, developed to automate the process of evaluating datasets and algorithms via IRT.
  
\end{itemize}

This paper is further organized as follows: Section 2 contextualizes the main issues covered herein, more precisely those about classical performance metrics, concept of innate ability, OpenML, Item Response Theory and the Glicko-2 system. Section 3 presents the related work and compares it with the present work. Section 4 presents the methodology that was used, explains how decodIRT and the Glicko-2 system were used. Section 5 discusses the results obtained. Section 6 presents the final considerations of the work and also brings a self-criticism made by the authors.

\section{Background}

\subsection{Classifier Evaluation and Benchmarks}

 Experiments in ML usually rely on robust benchmarks of datasets for algorithm training and testing. In fact, the creation of appropriate benchmarks are a key part of the research in ML. They are important pieces for the standardization of studies in the area, thus enabling the community to follow the progress over time, to identify which problems are still a challenge and which algorithms are best for certain applications. The lack of standardized benchmarks available results in many studies using their own sets of pre-processed datasets in their own way. This condition makes it difficult to compare and reproduce the results obtained by these studies \citep{bischl2017openml}.

\subsection{Classifier Evaluation}

In ML it is not enough just to train an algorithm, generate a model and start using it. It is very important to know whether the model that was generated was really able to learn to classify correctly. For this, the most common performance evaluation metrics can be applied. According to Kubat (2017) \cite{kubat2017introduction} there are different performance metrics and each one can be more interesting than the other depending on the aspect you want to evaluate.

Accuracy and error rate is one of the mostly used classic metrics. However, the result of a single performance metric can be misleading and not correctly reflect the true capability of a classifier \cite{kubat2017introduction}. In  Ferri, C., Hernández-Orallo, J. and Modroiu, R. (2009) \cite{ferri2009experimental} the authors experimentally analyzed the behavior of a total of 18 performance metrics. In the research, it is reinforced that the different performance metrics can generate different evaluations about the model's ability depending on the situation, that is, it depends on the data set being used. For example, in situations where there is an imbalance of classes or the dataset has few instances, a given metric may be preferable over the others. Thus, it is important to choose one or more specific metrics that are suitable to evaluate the model, taking into account the inner data complexities of the experiment.

\subsubsection{Innate ability}

According to Domingos (2012) \cite{domingos2012few}, the main objective of ML is generalization, that is, the algorithm that best manages to generalize during training is the best one to be chosen. Making an analogy with human life, the preferred algorithm would be one that has the best ``innate ability'' for learning. Given this situation, the following questions arise: Is it possible to use classic metrics to measure the innate ability of models? Which metric or metrics would be best suited to assess a model's ability? For this, first, it is important to define what the innate ability would be.

The innate can be conceptualized as: ``born with''. This means that the innate ability would then be an ability that is already present from the individual's birth. For example, people from a young age have immense talent in a given activity. When translating this concept into the field of supervised learning, the skill of a model would be its ability to learn to classify well, as this is the only activity to be performed. The innate ability would then be the ability of the algorithm to be able to classify well independently of the hyperparameters and datasets being used, as this would be a natural ability that the algorithm was ``born with''.

Classic metrics are aimed at evaluating the classifier facing a specific task, that is, whether the classifier can classify a dataset well. This approach only assesses the ability of the algorithm against a single dataset, which does not allow for defining whether the model would perform well in a general context, a situation that is sought to be explored with the concept of innate ability.

\subsection{OpenML}

OpenML is a repository that works as a collaborative environment where ML researchers can automatically share detailed data and organize tham to work more efficiently and collaborate on a global scale \citep{vanschoren2014openml}. It also allows ML tasks to be executed with the repository datasets using the preference algorithm and then share the results obtained within the platform, which minimizes the double effort. In addition, OpenML also makes it possible for new datasets to be made available by users, thus challenging the community to run algorithms on the dataset by using specific parameters to solve a given ML task (e.g., classification) \citep{vanschoren2014openml}. 

The platform can be divided into four main classes, namely: Datasets, Tasks, Flows and Runs. In the Datasets class, the existing datasets in OpenML are provided. Tasks describe what to do with the dataset, define what types of inputs are provided, what types of outputs should be returned, and the scientific protocols that can be used. Flows are precisely the learning algorithms that are applied to solve Tasks. While Runs is the application of a given Flow to a given Task \citep{vanschoren2014openml}.

In addition to the four main classes, OpenML also has the Study class which allows for combining the four main classes into studies to share with the online community or simply to keep a record of a work \citep{openml}. The Study class also allows for the creation of benchmark suites that can be translated as a set of tasks that are selected to evaluate algorithms under specific conditions. It creates benchmarks that enable the experiments performed on them to be clearly reproducible, interpretable and comparable \citep{bischl2017openml}.

\subsection{OpenMLCC-18 benchmark}

The creation of appropriate benchmarks are a key part of the research in ML. They are important pieces for the standardization of studies in the area, thus enabling the community to follow the progress over time, to identify which problems are still a challenge and which algorithms are best for certain applications. The lack of standardized benchmarks available results in many studies using their own sets of pre-processed datasets in their own way. This condition makes it difficult to compare and reproduce the results achieved by these studies \citep{bischl2017openml}.

In this context, OpenML also has the advantage of providing several reference benchmarks, such as the OpenMLCC-18 \footnote{Link to access OpenML-CC18: \url{https://www.openml.org/s/99}}. Proposed by Bischl (2017) \cite{bischl2017openml}, OpenML-CC18 is a classification benchmark composed of 72 existing OpenML datasets from mid-2018 and which is aimed at addressing a series of requirements to create a complete reference set. In addition, it includes several datasets frequently used in benchmarks published in recent years.

According to Bischl \cite{bischl2017openml}, the properties used to filter the datasets are: (a) Number of instances between 500 and 100,000; (b) Number of features up to 5000; (c) At least two classes targeted, where no class has less than 20 instances in total; (d) The proportion between minority and majority classes must be above 0.05; (e) Datasets cannot have been artificially generated; (f) Datasets must allow for randomization through a 10-field cross-validation; (g) No dataset can be a subset of another larger dataset; (h) All datasets must have some source or reference available; (i) No dataset should be perfectly classifiable by a single feature; (j) No dataset should allow a decision tree to achieve 100\% accuracy in a 10-field cross-validation task; (k) Datasets cannot have more than 5000 features after a \textit{one-hot-encoding} process on categorical features; (l) The datsets cannot have been created by binarizing regression or multiclass tasks; (m) No dataset can be sparse.

Therefore, it is understood that OpenML has a lot to contribute to research in the field of machine learning. In the previous work \citep{cardoso2020decoding}, an initial analysis of OpenML-CC18 was performed using IRT, which allowed for the generation of new relevant metadata about the complexity and quality of the benchmark, such as the difficulty and discriminative power of the data. This research seeks to deepen this analysis by looking for a subset of datasets within OpenML-CC18 that is as good or perhaps better than the original, while using IRT to find a more efficient benchmark subset that maintains the characteristics of the original.

\subsection{Item Response Theory}

According to de Andrade, Tavares and da Cunha Valle (2000) \cite{de2000teoria}, to assess the performance of individuals in a test, traditionally, the total number of correct answers is used to rank the individuals being evaluated. Despite being common, this approach has limitations to assess the actual ability of an individual. On the other hand, IRT allows for the assessment of latent characteristics of an individual that cannot be directly observed and it is aimed at presenting the relationship between the probability of an individual correctly responding to an item and their latent traits, that is, their ability in the assessed knowledge area. One of the main characteristics of the IRT is to have the items as central elements and not the test as a whole; the performance of an individual is evaluated based on their ability to hit certain items of a test and not how many items they hit.

Also according to \cite{de2000teoria}, the IRT is a set of mathematical models that seek to represent the probability of an individual to correctly answer an item based on the item parameters and the respondent's ability, where the greater the individual's ability, the greater the chance of success. The various proposed models depend fundamentally on three factors, namely:

\begin{enumerate}
  \item The nature of the item: whether it is dichotomous, in which only whether the answer is right or wrong is considered. Or whether it is non-dichotomous, where more possible answers are considered.
  \item Number of populations involved, whether it is just one or more than one.
  \item Amount of latent traces being measured.
\end{enumerate}

Logistic models for dichotomous items are the most used ones. For these items, there are basically three types of models, which differ by the number of item parameters being used. These are known as 1, 2 and 3 parameter logistic models. The 3-parameter logistic model, called 3PL, is the most complete model of the three, where the probability of an individual $j$ correctly answering an item $i$, given their ability, is defined by the following equation: 



\begin{equation}
    P(U_{ij} = 1\vert\theta_{j}) = c_{i} + (1 - c_{i})\frac{1}{1+ e^{-a_{i}(\theta_{j}-b_{i})}}
\end{equation}

\noindent Where:

\begin{itemize}
\setlength\itemsep{.25cm}
  
  \item $U_ {ij}$ is the dichotomous response that can take the values 1 or 0, being 1 when the individual \textit{j} hits the item \textit{i} and 0 when he misses it;
  
  \item $\theta_{j}$ is the ability of the individual \textit{j};
  
  
  \item $b_{i}$ is the item's difficulty parameter and indicates the location of the logistic curve;
  
  \item $a_{i}$ is the item's discrimination parameter, i.e., how much the item \textit{i} differentiates between good and bad respondents. This parameter indicates the slope of the logistic curve. The higher its value, the more discriminating the item is;
  
  \item$c_{i}$ is the guessing parameter, representing the probability of a casual hit. It is the probability that a respondent with low ability hits the item.

\end{itemize}

Although theoretically the discrimination parameter can vary from $-\infty$ to $+\infty$, negative discriminatory values are not expected, as this means that the probability of success is greater for individuals with lower ability values, which goes against what is expected by the IRT \cite{de2000teoria}. The other two logistic models can be obtained by simplifying the 3PL. For 2PL the guessing parameter is removed, i.e., $c_{i}=0$. The guessing parameter also represents the lowest possible correct answer probability. So, removing it eliminates the possibility that a low-ability respondent might respond correctly by luck. For 1PL the discrimination parameter is also removed, assuming that $a_{i}=1$. The discrimination parameter concerns the quality of the item itself, where negative discrimination can alert about a possible inconsistency; so, when removing it, useful information about the item's integrity can be lost.

To estimate the item parameters, the response set of all individuals for all items that will be evaluated is used. Unlike the classic assessment approach, the IRT is not designed to generate a final respondent score. Its purpose is to provide a ``magnifying glass'' that allows for observing the individual's performance more specifically on each item and estimate a likely ability level in the area under assessment. However, when taking a test, it is common to wait for a final score. Therefore, the IRT also has the concept of True-Score \cite{lord1984comparison}, which is the sum of the correct probabilities calculated for each item in the test. The True-Score is then used to set a final score that summarizes the respondent's performance. Based on the above, it is understood that IRT can be an adequate approach to assess the real ability of classifiers and the complexity of datasets.

\subsection{Glicko-2 System}

Although IRT already has the True-Score calculation as its own metric to generate a final score, it is understood that, in order to properly explore the concept of classifiers' ability, it is necessary to apply a more robust evaluation method together with the IRT. Given this, the herein work proposes the use of rating systems to summarize the data generated by the IRT and to define a final score that is capable of measuring the classifiers ability, given the fact that rating systems are widely used to measure an individual's ability in an activity, where rating is the numerical value that measures the ability \cite{vevcek2014chess}.

These systems are usually used in competitions to measure the ``strength'' of competitors, where each individual will have their own rating value and after a match this value is updated depending on the result (win, draw or defeat). Among the existing rating systems, Glicko-2 is the update of the Glicko system developed by Mark E. Glickman \cite{glickman2012example} to measure the strength of chess players. The Glicko system was developed in order to improve the \textit{Elo} system \cite{elo1978rating} taking into account the players' activity period to ensure greater reliability to the rating value \cite{samothrakis2014predicting}.

In the Glicko-2 system, each individual has three variables that are used to measure the statistical strength, namely: the rating value R, the rating deviation (RD) and the volatility ($\sigma$). Despite being very approximate, it cannot be said that the rating value perfectly measures an individual's ability, as it is understood that this value may undergo some variation. For this, the Glicko system has the RD, which allows for calculating a 95\% reliable range of rating variation, using the formula: $[R-2RD, R+2RD]$. This means that there is a 95\% chance that the individual's actual strength is within the calculated range. Therefore, the smaller the RD value, the higher the rating precision \cite{glickman2012example,samothrakis2014predicting}.

To measure how much fluctuation the rating is within its RD range, Glicko uses volatility. Thus, the higher the volatility value, the greater the chances of the rating having large fluctuations within its range; and the lower the volatility, the more reliable the rating is. For example, in a dispute between individuals with low volatility values, based on their ratings, it is possible to state more precisely who is the strongest \cite{samothrakis2014predicting,vevcek2014chess}.

The Glicko-2 system uses the concept of rating period to estimate rating values, which consist of a sequence of matches played by the individual. At the end of this sequence, the Glicko system updates the player's parameters by using the opponents' rating and RD along with the results of each game (e.g., 1 point for victory and 0 for defeat). If the individual is being evaluated for the first time, the Glicko system uses standardized initial values, namely: 1500 for rating, 350 for RD and 0.06 for volatility \cite{glickman2012example}.

\section{Related works}

\subsection{ML meets IRT}

Since it is a recent approach, there are few works aimed at applying IRT in studies involving the fields of AI. Prudêncio et al. (2015) \cite{prudencio2015analysis} seek to take the first steps to employ IRT in ML. The purpose of this research is to understand the relationship between a dataset considered difficult and the performance of the models. Where they consider that once they get the knowledge that a given classifier performs better in datasets with instances considered difficult, this makes this method preferred over the others. This analysis is compared to the methodology used for psychometric analysis of the proficiency level of students on a test using the IRT.

In this study, several Random Forests models with different numbers of trees were used to generate the set of responses to estimate the item parameters. For a case study, the Heart-Statlog dataset and the two-parameter logistic model (2PL) were used, focusing on the difficulty parameter. In addition, the work also uses IRT to identify instances considered noise through the intentional insertion of false instances. To compare the performance of classifiers, from the calculation of the hit probability, three different classifiers were used: Naive Bayes, Logistic Regression and Random Forests.

Another work that also employs IRT in ML is Martínez-Plumed et al. (2016) \cite{martinez2016making}. Their objective is also to apply IRT as a method to understand how different classification algorithms behave when faced with difficult instances of a dataset. In addition to trying to verify whether the so-called difficult instances are actually more difficult than the others or whether they are just noise. Furthermore, it also seeks to provide an overview of IRT and how it can be used to resolve the many existing issues about machine learning.

This work is a continuation of the work mentioned above, its main differences involve the use of several classifiers from 15 families of algorithms to generate the set of answers. As a case study, they use the Cassini and Heart-Statlog datasets. In addition to proposing the use of artificial classifiers to serve as a baseline between optimal and bad classification in a linear way. This time, the three-parameter logistic model (3PL) was chosen. In addition to presenting the concept of Classifier Characteristic Curve (CCC) as a way to visualize and analyze the variation in the classifiers' performance on instances with different values of difficulty and discrimination.

Martínez-Plumed et al. (2019) \cite{martinez2019item} is the most complete work, as it aims to describe a pipeline of how to apply IRT in machine learning experiments and it explores the advantages of its use with a focus on supervised learning. In the work, the authors discuss how each item parameter can be used to carry out a more in-depth analysis about the result of the classifiers. Also, the difference in the use of different logistic models of the IRT is observed, whereby the 3PL presents the most consistent results.

To perform the IRT analyses, this study used a set of 12 real datasets plus an artificial dataset. In addition, 128 classifiers from 11 different algorithm families were used. The objective is to explore why instances have different item parameter values and how this affects the performance of various learning algorithms. At the end of the work, the authors also suggest five main areas of ML in which IRT can be applied, namely: using IRT to improve classifiers; creation of portfolios of algorithms; classifier selection; improve understanding of the complexity of datasets; and assessing classifiers using IRT.

Like this paper, Martinez-Plumed and Hernandez-Orallo (2018) \cite{martinez2018dual} use the IRT to assess benchmarks according to the difficulty and discrimination estimators; but, unlike the other works mentioned above, the focus is on reinforcement learning instead of supervised learning. The authors use the benchmarks Arcade Learning Environment (ALE) \cite{bellemare2013arcade} and General Video Game AI (GVGAI) \cite{perez20152014}, which are remarkable benchmarks that allow for observing the performance of AI agents in different problems. In the herein study, dual indicators are proposed to evaluate both the benchmarks and the AI agents - stemming from different reinforcement learning algorithms, using the IRT concepts and proposing the generality indicator. This can be understood as the ability of an agent to solve all tasks up to a certain level of difficulty, which can be limited by available resources.

The authors apply the IRT 2PL logistic model for dichotomous items. For this, they use the human performance threshold in each analyzed game, where: if the AI agent's performance can equal or surpass human perfoamnce, then it is considered the correct answer; otherwise, it is an incorrect answer. In addition to benchmarking, the authors also use IRT's estimated ability and generality to assess agents in order to use IRT to create the most suitable benchmarks by selecting games with high discrimination values and accurately measuring whether the AI agent is really generalizing or is specializing in specific tasks.



\subsection{Benchmarking}

In addition to OpenML-CC18, provided by OpenML, other works also highlight the importance of creating and maintaining good benchmarks, such as Nie et al. (2019) \cite{nie2019adversarial}. In the herein paper, the authors propose a new benchmark for NLI (Natural Language Inference), whereby the benchmark is developed through iterative human-and-model-in-the-loop adversary procedure. In this format, humans first write problems that models cannot classify correctly. The resulting new hard instances serve to reveal model weaknesses and can be added to the training set to create stronger models. Therefore, the new model undergoes the same procedure to collect weaknesses in several rounds, where after each cycle a new stronger model is trained and a new set of tests is created. This cycle can be repeated endlessly to create stronger models and harder benchmarks after each iteration.

Based on this premise, Facebook launched Dynabench \cite{facebook}, a platform for dynamic data collection and benchmarking. The goal is to use the adversary method to iteratively create SOTA (state of the art) models and benchmarks, so you can create a benchmark that doesn't get outdated over time.


Like the studies presented above, \cite{prudencio2015analysis,martinez2016making,martinez2019item}, this work also seeks to use IRT as a tool for analyzing datasets and classifiers. One of the objectives of this research is to evaluate the well-known benchmark OpenML-CC18 according to the viewpoint of IRT, in order to explore its evaluation ability. Also proposede is using the Glicko-2 rating system in conjunction with IRT as a new strategy to perform a more robust assessment of a classifier's strength and to assess the quality and efficiency of the subsets of a benchmark. The generality metric proposed by \cite{martinez2018dual} can be compared to the concept of innate ability explored herein, as well as the use of the discrimination parameter to filter and to choose what games would be more suitable to compose a specific benchmark is similar to the strategy adopted in this work to create more efficient benchmarks. And like Dynabench \citep{nie2019adversarial}, this work is aimed at creating and maintaining quality benchmarks, thus evaluating their ability to test classifiers through IRT parameters.


Additionally, the decodIRT is presented, which allows for automating the process of analyzing classifiers and datasets through IRT. Where datasets are automatically downloaded from the OpenML platform, the user only needs to choose which dataset they would like to evaluate. The goal is to refine the benchmark choice through IRT, looking for the most robust and difficult set of datasets available in OpenML.

\section{Materials and methods}

\begin{figure}[htbp]
\centering
\includegraphics[width=1 \textwidth]{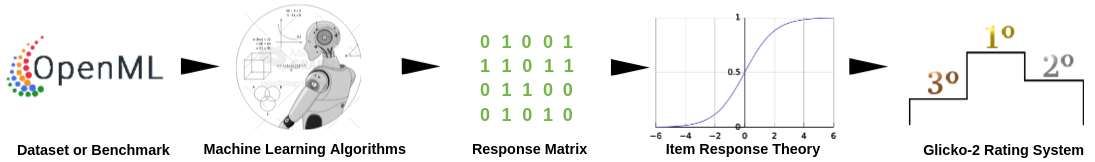}
\caption{Flowchart of the proposed methodology.}
\label{fig0}
\end{figure}

Figure \ref{fig0} illustrates the methodology proposed for applying the TRI and the Glicko-2 system to evaluate ML models and benchmarks:

\begin{enumerate}
    \item An OpenML supervised learning dataset or benchmark is chosen. Data is split into training and testing;
    \item Several ML models/classifiers are generated that will be trained and tested using datasets;
    \item The answer from these classifiers is collected in matrix form. Each row is a classifier, each column is a test instance, and the value represents whether the instance was correctly classified or not. There is a response matrix for each dataset;
    \item The answer matrix is used to calculate the item parameters of the test instances of each dataset. Subsequently, the probability of correct answer and the True-score of each classifier for each dataset are calculated;
    \item Using True-score, calculate the Glicko ratings for each classifier, with each dataset being taken as a Glicko rating period.
\end{enumerate}

\subsection{Fitting an IRT model}

Although generally applied for educational purposes, IRT has recently been extended to AI and more specifically to ML \cite{prudencio2015analysis,martinez2016making,martinez2019item}. For this, the following analogy is used: the datasets are the tests, the instances of a dataset are the items and the classifiers are the respondents. 

Despite the existence of three IRT logistic models, the 3PL was used for this research work, considered by \citep{martinez2019item} as the most complete and consistent logistic model. Furthermore, as stated in previous sections, for the other two logistic models the guessing is always considered equal to 0. However, it is not possible to state that a model cannot correctly classify an instance by luck. Therefore, the great importance of maintaining the guessing parameter to calculate the probability of correctness of the models is understood.

The item parameters are then used to evaluate the datasets directly by reporting the percentage of difficult instances with great discriminative power and with a great chance of random hits. Therefore, it is possible to have a view of the complexity of the evaluated datasets and how different classifiers behave in the challenge of classifying different datasets.

To calculate the probability of correct answer, one must first estimate the item parameters and the ability of respondents. According to \cite{martinez2016making}, there are three possible situations. In the first, only the item parameters are known. In the second situation, only the ability of the respondents is known. And in the third, and also the most common case, both the items parameters and the respondents ability are unknown. This paper work lies in the third case and for this situation, the following two-step interactive method proposed by Birnbaum \cite{birnbaum1968statistical} is applied:

\begin{itemize}
  \item At first, the parameters of each item are calculated only with the answers of each individual. Initial respondent capability values can be the number of correct answers obtained. For classifiers, this study used the accuracy obtained as the initial ability.
  
  \item Once obtained the item parameters, the ability of individuals can be estimated. For both item parameters and respondent ability, simple estimation techniques can be used, such as maximum likelihood estimation \cite{martinez2016making}.
\end{itemize}

\subsection{decodIRT tool}

To build the IRT logistic models and to analyze the benchmarks, the decodIRT \footnote {Link to the source code: \url{https://github.com/LucasFerraroCardoso/IRT_OpenML}} tool initially presented in \cite{cardoso2020decoding} was used. The main objective of the DecodIRT is to automate the analysis of existing datasets in the OpenML platform as well as the proficiency of different classifiers. For this, it depends on the probability of correct answer derived from the logistic model of IRT and the item parameters and the ability of respondents.

\begin{figure}[htbp]
\centering
\includegraphics[width=.75 \textwidth]{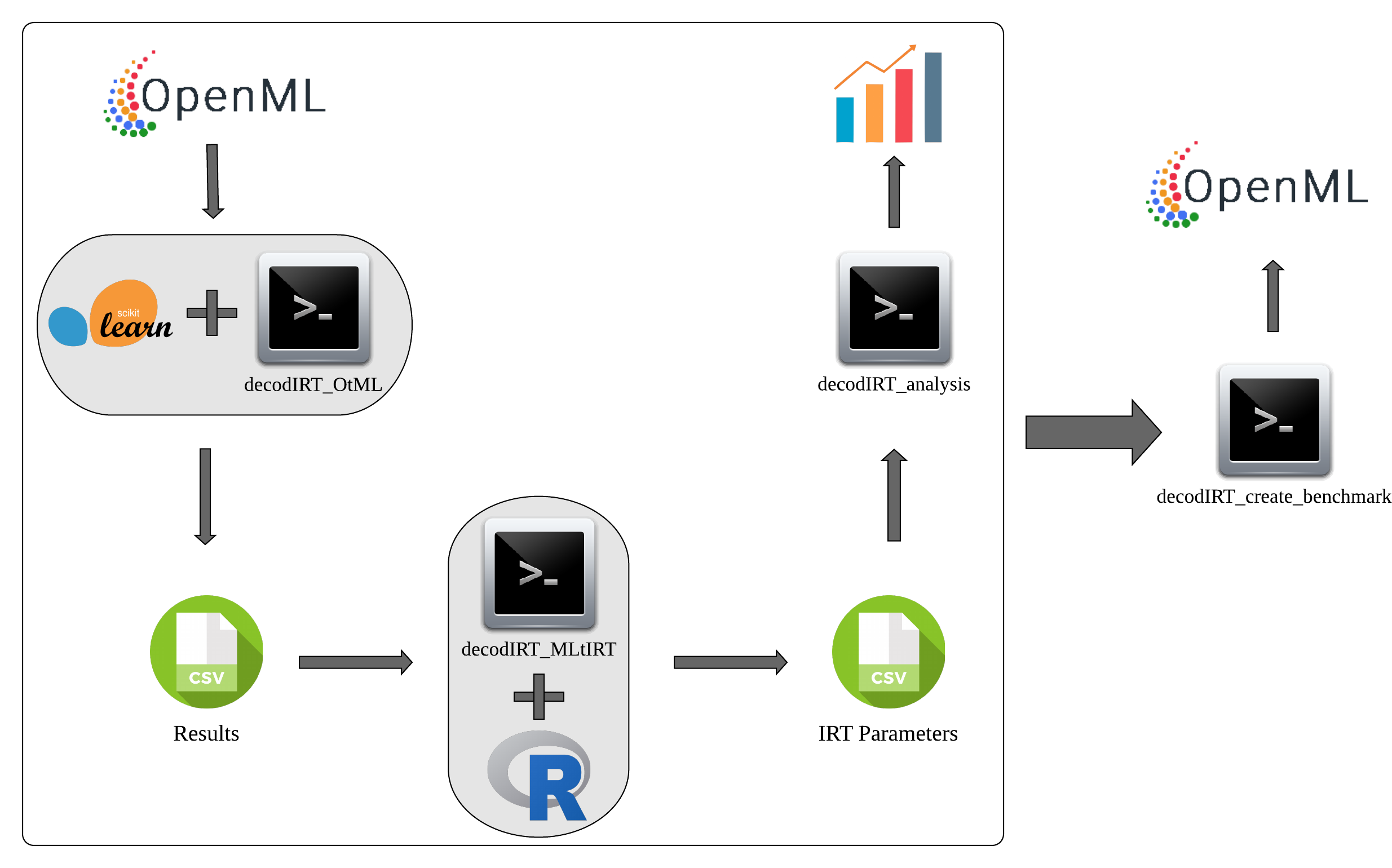}
\caption{Flowchart of the decodIRT pipeline.}
\label{fig1}
\end{figure}

As can be seen in Figure \ref{fig1}, the decodIRT tool consists of a total of four steps (scripts), with three main scripts (within the square) designed to be used in sequence. The first script is responsible for downloading the OpenML datasets, generating the ML models and placing them to classify the datasets. Then, a response matrix is generated, which contains the classification result of all classifiers for each test instance. The response matrix is the input to the second script, which in turn is responsible for calculating the item's parameters. The last script will use the data generated by the previous scripts to rank the datasets using the item parameters and to estimate the ability, calculate the response probability and the True-Score of each model.

DecodIRT was modified to work as a package too, where the fourth script (outside the square) was developed to facilitate the use of the tool by the user. Running the tool automatically, and allowing the user to use IRT estimators to evaluate benchmarks, creates benchmark subsets and store them directly in OpenML.

\subsection{Step 1: decodIRT\_OtML}

The first script is intended for downloading selected datasets from OpenML and running the models on the datasets to get the answers that are used to estimate the item parameters. As usual, the datasets are divided into a training set and a test set. So, the answers from the classifiers are obtained only for the test set. By definition, a stratified split of 70/30 is performed; but, for very large datasets, the split is handled so that the test set is composed of 500 instances at the most. This number of instances is considered large enough for analysis and will be better justified later on.

All ML models are generated by using Scikit-learn \cite{pedregosa2011scikit} as a standard library. Three different sets of classifiers are generated. The first set is composed only of Neural Network (MLP) models, amounting to 120 MLP models, where the depth of the networks gradually increases from 1 to 120. The second set is composed of 12 classifiers from different families that are evaluated herein, namely: Naive Bayes Gaussian standard, Naive Bayes Bernoulli standard, KNN of 2 neighbors, KNN of 3 neighbors, KNN of 5 neighbors, KNN of 8 neighbors, Standard Decision Trees, Random Forests (RF) with 3 trees, Random Forests with 5 trees, Standard Random Forests, Standard SVM and Standard MLP. The models classified as standard mean that the standard Scikit-learn hyperparameters were used. All models are trained through 10-field cross-validation.

The third set of models is composed of 7 artificial classifiers. The concept of artificial classifiers is initially presented in \cite{martinez2016making}, as follows: an optimal classifier (gets all the classifications right), a pessimal one (all misses), a majority (classifies all instances with the majority class), a minority (classifies with the minority class) and three random classifiers (sort randomly). This set is used to provide performance threshold indicators for real classifiers. And despite using OpenML as the base repository, decodIRT also allows the user to use local datasets and define training and testing sets specifically.

\subsection{Step 2: decodIRT\_MLtIRT}

This script's function is to use the responses generated by the classifiers and estimate the item parameters for the test instances. As stated previously, the logistic model for dichotomous items is used, which means that regardless of the number of classes existing in each dataset, it is only considered if the classifier was right or wrong in the classification of each instance.

To calculate the item parameters, the Ltm package \cite{rizopoulos2006ltm} for the R language is used, which implements a framework containing several mechanisms for the calculation and analysis of the IRT. The Rpy2 package \cite{gautier2008rpy2} was used to perform Python communication with the R packages. As mentioned previously, the maximum limit of 500 instances for estimating item parameters was defined. According to \cite{martinez2019item}, packages that estimate the IRT item parameters may get stuck in a local minimum or not converge if the number of items is too large. This is not strange. As the IRT is used for psychometric tests, it is very unusual for these tests to have such a large number of questions. Thus, it is recommended that less than 1000 instances be used to estimate the parameters.

\subsection{Step 3: decodIRT\_analysis}

The third script of the main set is responsible for performing the analysis and organizing the data generated by the previous scripts to render the data easier to read. Among the various functions of this script is the creation of dataset rankings by item parameter. Each ranking will organize the datasets according to the percentage of instances with high values for each of the parameters, i.e. values above a certain threshold. For example, the difficulty ranking will sort the datasets by the number of instances with high difficulty values. Limits can be defined by the user.

In case the user does not define any specific limit, default limit values are used, as based on \cite{adedoyin2013using}. In the mentioned work, the authors point out that, for an item to be considered difficult, the value of its difficulty parameter must be greater than 1. Very discriminative items have a discrimination value above 0.75. And for guessing, the limit value is 0.2. Analyzing the percentages of item parameters is one of the interests of this work.

Before calculating the classifiers correct answer probability for the instances, one must first estimate the ability of the classifiers, as explained in Birnbaum's method. Both to estimate the ability $\theta$ and to calculate the probability of correct answer use the Catsim package \cite{meneghetti2017application} from Python. For this, the instances are sorted according to their difficulty and divided into 10 groups; then, they are used in ascending order to estimate the ability of the classifiers. After this step, the probability of correct answer can then be calculated.

In addition, this script also implements the True-Score \citep{lord1984comparison} concept explained earlier in order to score the performance of classifiers. The True-Score, then, is also used as input for the generation of the rating values of the Glicko-2 system that is used to evaluate the performance and the innate ability of the classifiers.

\subsection{Step 4: decodIRT\_create\_benchmark}

This last script works as a benchmark builder through IRT. It allows the user to create new benchmark sets within the OpenML platform by using the item parameters to choose the best set of datasets. For this, the script uses the decodIRT tool as a library, where the user can choose from the OpenML the set of datasets he wants to evaluate with the IRT, which item parameter he wants to use and the cut percentage. For example, the user may select the difficulty parameter with a 20\% cur percentage. This means that the new benchmark will be composed of the 20\% most difficult datasets from the original set.

To add new benchmarks in OpenML, the platform's Study class is used. This class allows users to create complete studies involving ML, ranging from the set of datasets used to which algorithms and forms of assessment and training that were used \cite{openml_study}.

\subsection{Ranking of classifiers by the Glicko-2 system}

Due to the fact that rating systems are commonly used in competitions to apply the Glicko-2 \cite{glickman2012example} system to evaluate the classifiers, it was necessary to simulate a competition between them. The simulated competition is a round-robin tournament, where each classifier will face each other and at the end of the competition will create a ranking with the rating of the models.

The competition works like this: each dataset is seen as a classification period in the Glicko-2 system, so that all classifiers face off in each dataset. To define the winner of each contest, the True-Score values obtained by the models facing each other are used. This happens as follows: if the True-Score value is greater than the opponent's, it is counted as a victory; if the value is lower than that of the opponent, it is counted as a defeat; and if the values are equal, then it sets up a tie between the models. In addition, the Glicko system calls the result of a match to assign a score to the opponents. For this, the scoring system applied in official Chess competitions was used, where victory counts as 1 point, defeat as 0 point and draw counts as 0.5 point.

Thus, after each dataset, the rating RD and volatility values of the classifiers are updated and used as the initial value for the next dataset. Once all datasets are finalized, the final rating values are used to create the final ranking that will be used to evaluate the models.

\subsection{OpenML-CC18 datasets}

OpenML-CC18 was chosen to be the case study of this work, one of the main objectives of which being to evaluate benchmarks through the IRT standpoint in order to provide greater reliability in the use of this benchmark. This section will present the datasets that were selected from OpenML-CC18 to be evaluated using the decodIRT tool.

Despite having 72 datasets, only 60 were used in this work. This was for two main reasons:

\begin{enumerate}
    \item The size of the datasets, where 11 have more than 30,000 instances, were soon considered too large and would take a long time to run all decodIRT models;
    \item Could not generate item parameters for dataset ``Pc4''. R's Ltm package could not converge even using just under 500 test instances.
\end{enumerate}

Despite that, the final amount of datasets used still corresponds to 83.34\% of the original benchmark. All datasets evaluated are from tabular data and the characterization of the benchmark will be further explored in the next sections.

\subsection{Assessment of innate ability}

Given the definition of innate ability explained in the previous section, it is understood that its assessment can be done as follows:
\begin{enumerate}
    \item A benchmark composed of several datasets with different characteristics is defined. The chosen benchmark is OpenML-CC18 itself;
    \item A pool of classifiers composed of algorithms from the same family or from different families is assembled, always keeping the same configuration for each model. For this, the same set of real classifiers as decodIRT will be used;
    \item Test the different algorithms on the benchmark datasets. Step already performed by decodIRT;
    \item Different subsets of benchmarks are assembled from the original set. The subsets are assembled from the IRT estimators with decodIRT;
    \item For each subset of the benchmark, the rating ranking will be generated by the Glicko-2 system;
    \item The model with the best and most consistent performance is chosen as the one with the best innate ability.
\end{enumerate}

\section{Results and discussion}

The evaluation of the OpenML-CC18 benchmark through the IRT standpoint was done around the discrimination and difficulty parameters. It is understood that these parameters are directly linked to the data, in comparison with the guessing parameter, which is more linked to the performance of the respondents. The objective, then, is to evaluate the discriminatory power along with the difficulty of the datasets and later use them to evaluate the models' performance \footnote{All classification results can be obtained at \url{https://osf.io/wvptb/files/}}.

\subsection{Decoding OpenML-CC18 Benchmark}

\begin{figure}[htbp]
\centering
\includegraphics[width=.5 \textwidth]{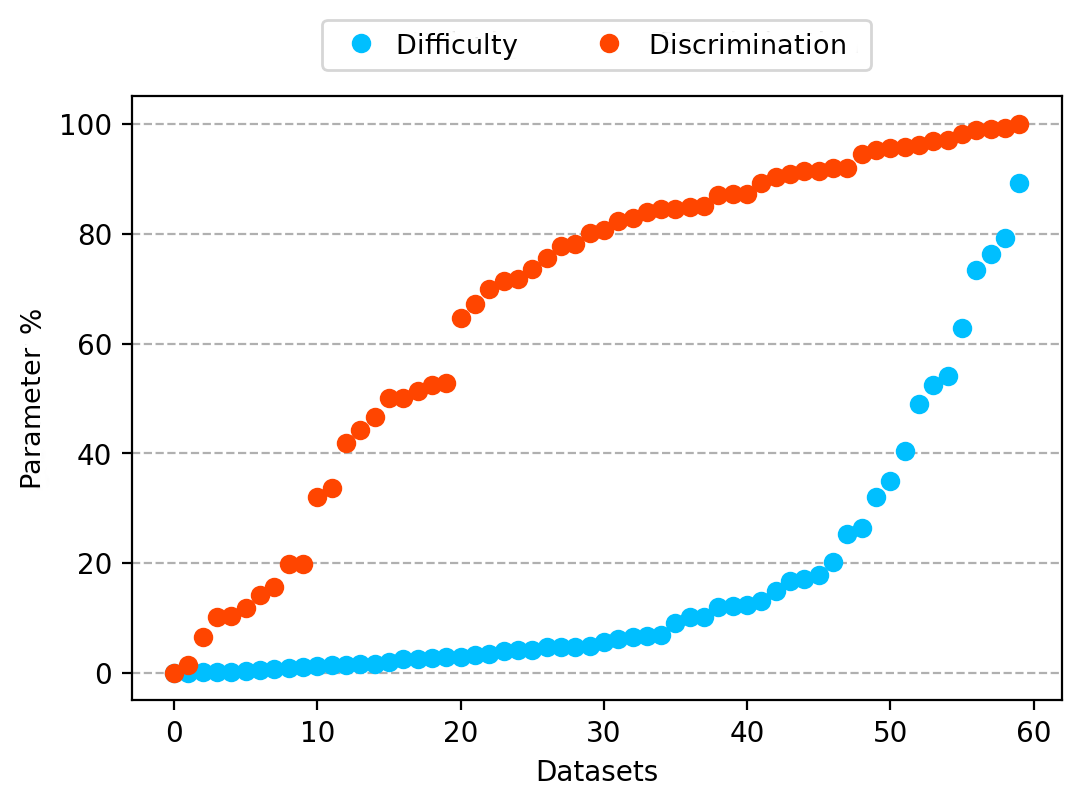}
\caption{The percentages of difficult and very discriminative instances arranged in ascending order. There is a certain percentage of discrimination and a percentage of difficulty that are in the same position on the X axis and do not necessarily correspond to the same dataset. ``tic-tac-toe'', ``credit-approval'' and ``optdigits'' are respectively the datsets with the most difficult instances. While ``banknote-authentication'', ``analcatdata\_authorship'' and ``texture'' are the most discriminative ones.}
\label{fig2}
\end{figure}


When looking at Figure \ref{fig2}, it is possible to notice an inversion relationship between the parameters of difficulty and discrimination. So, the rankings generated by the two parameters reveal that the most discriminating datasets are also the least difficult ones and vice-versa \footnote{Parameter rankings can be accessed at: \url{https://osf.io/jpygd/}}. This relationship is consistent with what is expected by the IRT, where it is normal that the easiest instances are good to differentiate the good from the bad classifiers, as it is thought that the more skilled classifiers will hit the easiest instances while the less skillful ones can make mistakes. Therefore, it is possible to affirm that the more difficult datasets are not adequate to separate the good and bad classifiers, despite being more challenging. Meanwhile, the easiest datasets are not suitable for testing the classification power of algorithms, but it allows for well differentiating the best from the worst.

\begin{figure}[htbp]
\centering
\includegraphics[width=.5 \textwidth]{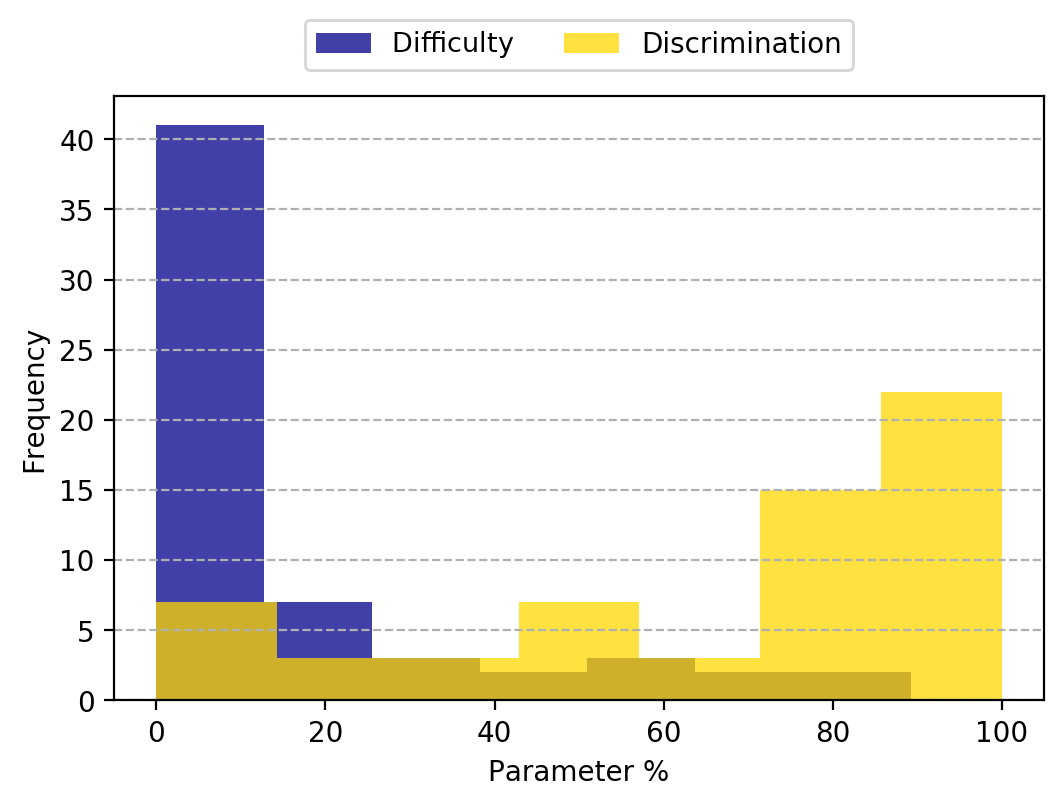}
\caption{Histogram of the number of datasets by the percentage of discriminative and difficult instances.}
\label{fig22}
\end{figure}

Among the evaluated datasets, only 7 are really challenging and have more than 50\% of difficult instances. While 49 of the total have less than 27\% of difficult instances. This means that only 11.67\% of the total evaluated datasets are hard while 81.67\% have more than 70\% easy instances. Therefore, the OpenML-CC18 benchmark should be used with caution and taking into account the purpose of its use. Figure \ref{fig22} strengthens the high discriminatory ability of the benchmark, where only 1/4 of the datasets have less than 50\% of low discriminatory instances and more than half of the total have at least 80\% of highly discriminatory instances.

Based on that, it is possible to infer that OpenML-CC18 is not considered as challenging as expected, but it has many appropriate datasets to differentiate the good and bad classifiers. In addition, item parameters allow the benchmark to be chosen more specifically. For example, if the objective is solely to test the algorithms' classification power, only the most difficult datasets can be used, disregarding testing with the entire benchmark.

\subsection{Classifiers performance on OpenML-CC18}

\begin{figure}[htbp]
\centering
\includegraphics[width=.7 \textwidth]{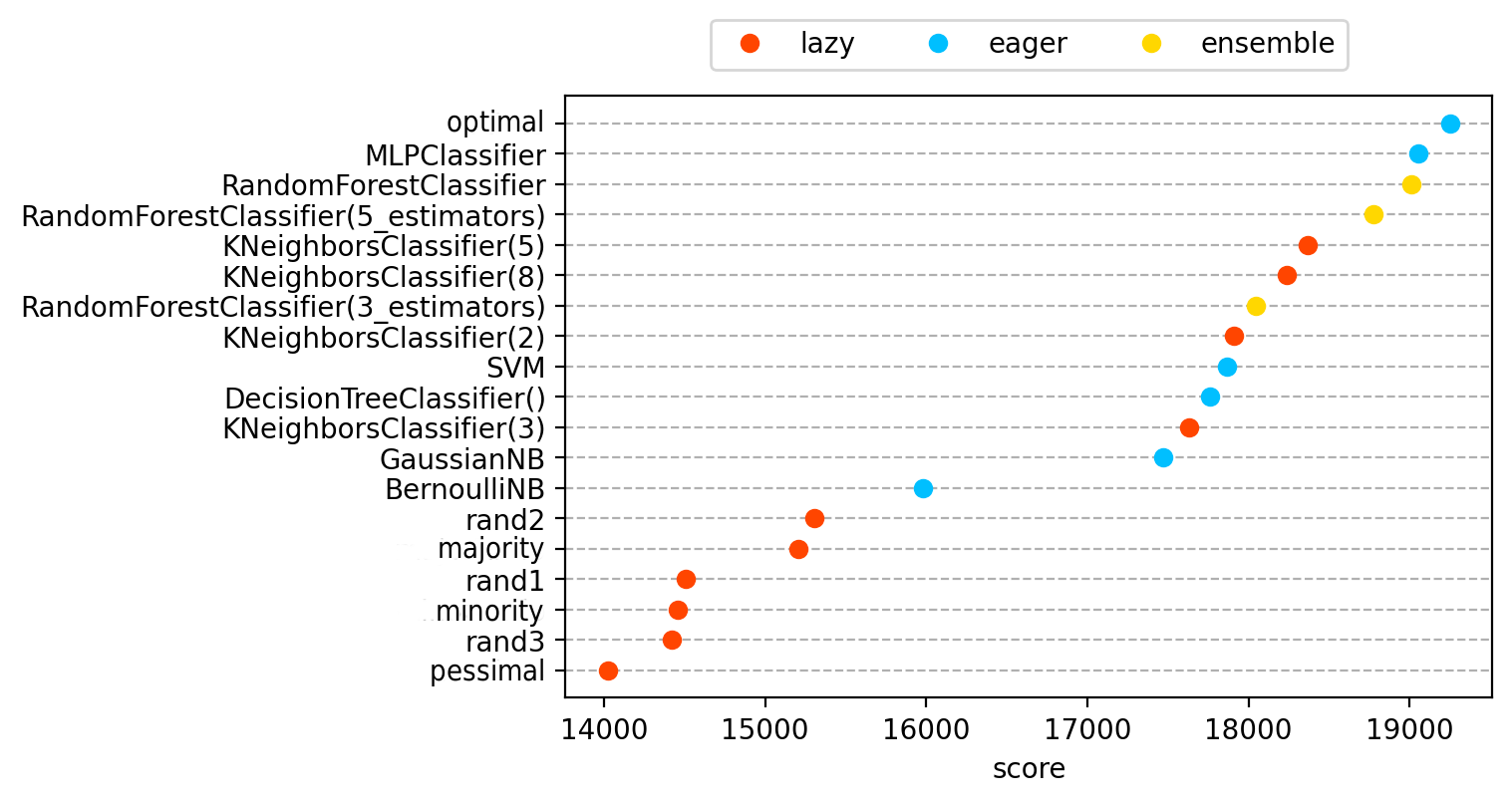}
\caption{Sum of all True-Scores obtained by the classifiers.}
\label{fig3}
\end{figure}

\begin{figure}[htbp]
\centering
\includegraphics[width=.7 \textwidth]{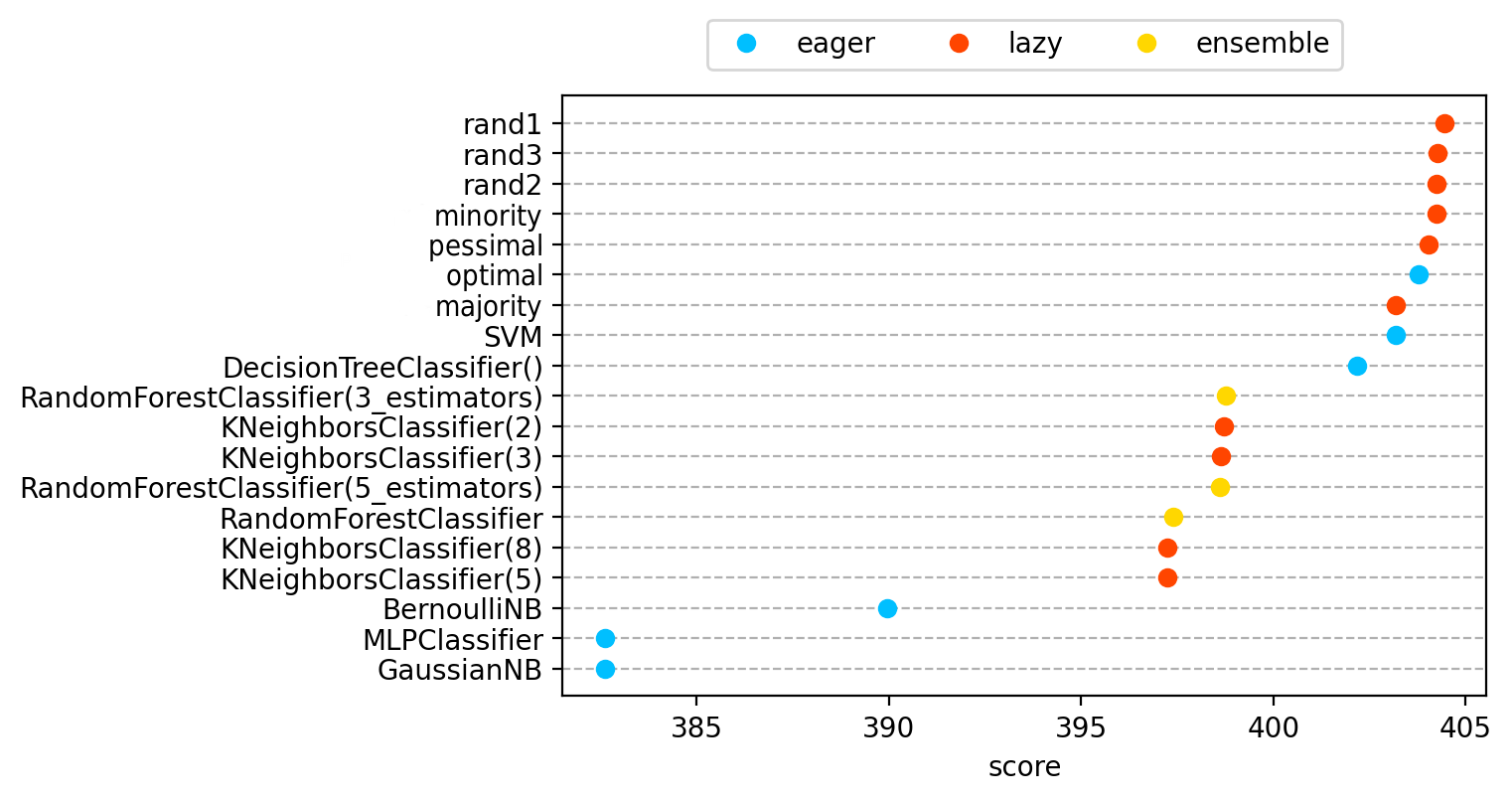}
\caption{The True-Score values obtained for the ``jm1'' dataset.}
\label{fig4}
\end{figure}

When taking into account only True-Score values 
obtained by the classifiers it is already possible to notice a pattern in their respective performances and create a ranking (refer to Figure \ref{fig3}). The artificial classifiers are observed to assume extreme positions in the ranking, as expected. For real classifiers, MLP takes the lead, but with a True-Score value very close to Random Forest. Although classifications similar to this one are the most common, in some specific cases the position of the classifiers is inverted and the worst models have the highest True-Score values, as can be seen in Figure \ref{fig4}.

This situation can happen due to the occurrence of many instances with negative values of discrimination. Since they are not expected by the IRT, negative values usually mean that there is something wrong with the item itself. For psychometric tests, this could mean a poorly formulated and ambiguous question, for example. When placing this concept in the ML field, negative discrimination may indicate some inconsistency in the instance, such as noise or outlier. Therefore, it can be inferred that datasets with many instances with negative discrimination may not be suitable for the formulation of a good benchmark. A possible future work would be to carefully analyze whether dataset characteristics are linked to these situations and how this can affect the performance of models. OpenML already has an extensive set of metadata about its datasets that can be used for this purpose.

\subsection{Classifiers performance by Glicko-2}

Although the True-Score values obtained already make it possible to observe and evaluate the performance of the classifiers, there is still a large amount of data to be evaluated individually before being able to point out which algorithm has the best innate ability. Therefore, the Glicko-2 system was applied in order to summarize the generated data and to identify the ``strongest'' algorithm. Table \ref{tab1} shows the final rating ranking that was obtained.

\begin{table}[h]
\begin{center}
\begin{minipage}{174pt}
\caption{Classifier rating ranking.}\label{tab1}%
\begin{tabular}{@{}llll@{}}
\toprule
Classifier & Rating & RD & Volatility\\
\midrule
optimal                                 & 1732.56 & 33.25 & 0.0603     \\ 
MLP                         & 1718.65 & 31.20 & 0.0617     \\ 
RandomForest                & 1626.60 & 30.33 & 0.0606     \\ 
RandomForest(5\_trees) & 1606.69 & 30.16 & 0.0621     \\ 
RandomForest(3\_trees) & 1575.26 & 30.41 & 0.0646     \\ 
DecisionTree              & 1571.46 & 31.16 & 0.0674     \\ 
SVM                                   & 1569.48 & 32.76 & 0.0772     \\ 
KNeighbors(3)               & 1554.15 & 30.74 & 0.0646     \\ 
GaussianNB                            & 1530.86 & 31.25 & 0.0683     \\ 
KNeighbors(2)               & 1528.41 & 30.40 & 0.0638     \\ 
KNeighbors(5)               & 1526.10 & 30.27 & 0.0630     \\ 
BernoulliNB                           & 1494.87 & 32.64 & 0.0770     \\ 
KNeighbors(8)               & 1457.78 & 30.25 & 0.0638     \\ 
minority                           & 1423.01 & 30.66 & 0.0631     \\ 
rand2                                 & 1374.78 & 30.27 & 0.0605     \\ 
rand3                                 & 1337.27 & 30.95 & 0.0600     \\ 
rand1                                 & 1326.38 & 31.42 & 0.0610     \\ 
majority                           & 1301.08 & 31.74 & 0.0666     \\ 
pessimal                               & 1270.46 & 31.74 & 0.0603     \\
\botrule
\end{tabular}
\footnotetext{Source: The author (2021).}
\end{minipage}
\end{center}
\end{table}

As in the True-Score ranking (refer to Figure \ref{fig3}) the position of the artificial classifiers is as expected. Optimal leads while the other artificial classifiers have ratings lower than all real classifiers, where it is also the MLP that has the highest rating among the real ones. However, MLP is closer to Optimal rating than RF is in third place. This situation differs from the True-Score ranking and from what was expected, as it was thought that Optimal would have a much higher rating than the others.

Despite the proximity of the rating of the MLP and Optimal be surprising, the strength of the MLP can be confirmed by observing the low volatility value. Overall, volatility is low for all classifiers, with caveats for SVM and Naive Bayes Bernoulli, which have the highest volatility values, respectively. This means that SVM and NB Bernoulli have the least reliable rating values of all, so they are more likely to vary widely within their respective RD ranges.

Furthermore, if you consider a high fluctuation in ratings within their RD ranges, the final ranking position may change sharply. For example, considering the largest negative change in MLP's RD, its new rating will be 1656.25. This would allow raters up to 4th place to be able to outperform the MLP should their ratings fluctuate as much upward as possible. However, for raters from 5th position onwards, no model could reach the MLP, even with the maximum fluctuation of their ratings as well.

Therefore, it is understood that there are groups of classifiers that have equivalent strength, while among the first three real classifiers it is not possible to say precisely which one is the strongest in the challenge proposed by OpenML-CC18. However, this situation also allows us to assume that the innate ability of MLP is better than that of algorithms below 4th position because, given the fact that tests were performed with several different datasets that have different IRT estimators, always keeping the model configuration, it can be assumed that the results obtained reflect the innate ability of the learning algorithms.

It is also important to point out that the optimization of models can have a fine-tuning effect on the decision limits of the classifiers, resulting in better performance in the most difficult datasets. However, this approach would not allow for a clean analysis of the models' innate ability.

\begin{figure}[htbp]
\centering
\includegraphics[width=0.7 \textwidth]{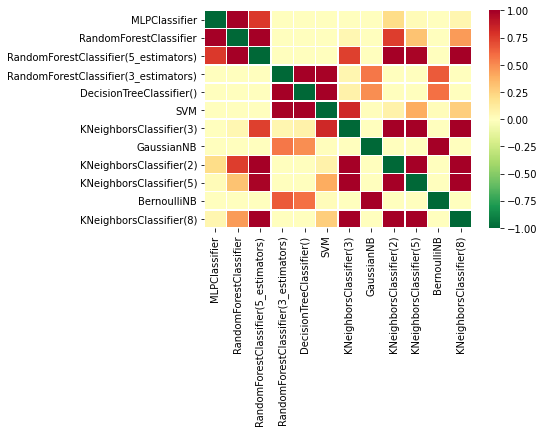}
\caption{Heatmap generated from the Nemenyi Test, using only the rating distributions of the real classifiers.}
\label{fig5}
\end{figure}

In order to provide the rating values that were generated with greater credibility, the Friedman test \cite{pereira2015overview} was performed. By that, the aim is to identify whether through the rating values, indeed, it is possible to differentiate the algorithms' innate ability. The Friedman test was calculated using only the rating values of the real classifiers, as they are the focus of the study. Its execution resulted in a p-value of approximately \num{9.36e-80}. The low p-value obtained means that, in fact, different distributions of ratings were found, which allowed the execution of the Nemenyi test \cite{nemenyi1962distribution}. The Nemenyi test is applied to identify which distributions differ from each other. Figure \ref{fig5} shows a Heatmap of the Nemenyi test.

When analyzing the Heatmap, it can be noted that the assumption of the existence of groups is proven, as the first three classifiers that have the highest ratings also have high p-value. This means that the Nemenyi test cannot differentiate them. And although they have the best performers, they all have a p-value for at least one lower performer. Other classifiers also have a high p-value, even though they are from different algorithm families. Therefore, it is not evident that there is a clear separation of classifiers from different groups. And this leads to believe that, although the Friedman test indicates the existence of distinct groups of classifiers, these differences are not statistically significant to indicate with certainty which algorithm is more skillful.

Based on the above, a deeper analysis of the performance of the algorithms is necessary in order to define which, in fact, has the greatest innate classification ability. For this, different subsets of OpenML-CC18, chosen from the estimators of the IRT, were generated and only then the algorithm that featured the best innate ability would be selected. Furthermore, it will also be explored whether there is one subset more efficient than the original within the generated benchmark subsets.

\subsection{Benchmarking evaluation with the aid of decodIRT}

According to Smith and Martinez (2014) \cite{smith2014reducing}, not all instances of a dataset are equally beneficial for learning. This means that some instances can help more than others in the model induction process, and it is even possible to have instances that hinder learning. By expanding this idea to a higher level, it is possible to figure out that in a benchmark, not all datasets are equally beneficial for evaluating models.

Given that, one can imagine that within a benchmark there may be a subset of datasets that is as good, or even better, than the original benchmark. Through this, it is also believed that decodIRT can help to find more efficient benchmarks. Due to using the item parameters of the IRT it is possible to select the datasets that have the highest percentage of instances with high values for each one of the item parameters.

According to the results presented above, it is known that the datasets occupy opposite positions in the Discrimination and Difficulty rankings. Therefore, it is understood that to generate a more efficient subset of datasets it is necessary to take into account the values of both parameters.

To assess whether there is a subset of OpenML-CC18 that is equally good or better than the original, three cutoff percentages were defined - 30\%, 50\% and 70\% - to select the most difficult and discriminating set of datasets using the decodeIRT. For example, for cutting 30\%, it will be the most discriminating 15\% datasets and the most difficult 15\% datasets. The assessment of the quality of the subsets is performed by analyzing the Glicko rating rankings generated after each percentage cut.


\begin{table}[h]
\begin{center}
\begin{minipage}{174pt}
\caption{Classifier rating ranking by benchmark subset 70\%.}\label{tab4}%
\begin{tabular}{@{}llll@{}}
\toprule
Classifier & Rating & RD & Volatility\\
\midrule
optimal                                 & 1689.83 & 31.59 & 0.0601     \\ 
RandomForest                         & 1672.18 & 30.67 & 0.0599     \\ 
MLP                & 1643.37 & 29.99 & 0.0604     \\ 
RandomForest(5\_trees) & 1628.19 & 30.17 & 0.0605     \\ 
RandomForest(3\_trees)                            & 1621.54 & 30.17 & 0.0632     \\ 
KNeighbors(8)               & 1604.03 & 30.17 & 0.0629     \\ 
KNeighbors(5)                                   & 1602.32 & 30.19 & 0.0638     \\
KNeighbors(2)              & 1580.18 & 29.90 & 0.0629     \\ 
SVM               & 1556.00 & 30.74 & 0.0677     \\
GaussianNB               & 1547.81 & 31.13 & 0.0684     \\  
DecisionTree                           & 1540.01 & 30.55 & 0.0661     \\ 
BernoulliNB               & 1526.66 & 31.45 & 0.0708     \\ 
KNeighbors(3) & 1519.88 & 29.81 & 0.0627     \\ 
majority                           & 1364.58 & 30.59 & 0.0637     \\ 
rand3                                 & 1323.52 & 30.45 & 0.0600     \\ 
rand2                                 & 1315.71 & 30.42 & 0.0604     \\
minority                                 & 1295.84 & 30.98 & 0.0611     \\ 
pessimal                               & 1291.72 & 31.00 & 0.0606     \\
rand1                           & 1287.40 & 31.11 & 0.0607     \\ 
\botrule
\end{tabular}
\footnotetext{Source: The author (2021).}
\end{minipage}
\end{center}
\end{table}

By looking at the ratings generated by the 70\% set (refer to Table \ref{tab4}), there are some variations in the positions of the real and artificial classifiers. In this new ranking, Random Forest took the lead over MLP, but it still has a rating value very close to the Optimal classifier, a situation that should not occur given the Optimal characteristics. Another point to consider is the proximity of the rating values in the 70\% set because, considering the maximum variation of the rantings in their respective RD intervals, the Optimal classifier can be reached by the KNeighbors(2) classifier, which is in the 8th position of the ranking. Such a condition makes it more difficult to try to separate the classifiers by their strength.

However, despite the proximity of ranking between the real classifiers and Optimal, the new benchmark set had a good positive point. The rating difference between the last real classifier and the first classifier, among the artificial ones that occupy the lower part of the table, has increased significantly. The original benchmark ranking value (refer to Table \ref{tab1}) has increased from 34.77 to 155.3, which makes real classifiers unreachable for artificial classifiers. And this situation is in line with what was expected, given the large differences in performance, and it probably occurs due to the greater discriminative ability of the datasets.


For the set of 50\%, the final performance of the classifiers and their ranking order is the closest to what was expected (refer to Table \ref{tab3}).

\begin{table}[h]
\begin{center}
\begin{minipage}{174pt}
\caption{Classifier rating ranking by benchmark subset 50\%.}\label{tab3}%
\begin{tabular}{@{}llll@{}}
\toprule
Classifier & Rating & RD & Volatility\\
\midrule
optimal                                 & 1724.29 & 32.78 & 0.0601     \\ 
RandomForest                         & 1678.31 & 31.21 & 0.0600     \\ 
MLP                & 1655.45 & 30.52 & 0.0602     \\ 
RandomForest(5\_trees) & 1644.98 & 30.84 & 0.0600     \\ 
RandomForest(3\_trees)                            & 1634.40 & 30.32 & 0.0618     \\ 
KNeighbors(2)              & 1620.72 & 30.33 & 0.0609     \\ 
KNeighbors(8)               & 1593.82 & 30.30 & 0.0621     \\ 
KNeighbors(5)                                   & 1591.09 & 30.31 & 0.0628     \\ 
KNeighbors(3) & 1569.81 & 30.09 & 0.0603     \\ 
BernoulliNB               & 1558.71 & 31.32 & 0.0672     \\ 
SVM               & 1549.54 & 31.14 & 0.0676     \\
DecisionTree                           & 1522.45 & 30.66 & 0.0646     \\ 
GaussianNB               & 1510.21 & 30.64 & 0.0635     \\ 
majority                           & 1377.13 & 30.63 & 0.0623     \\ 
rand3                                 & 1287.90 & 31.64 & 0.0600     \\ 
rand1                           & 1276.45 & 31.75 & 0.0602     \\ 
rand2                                 & 1274.34 & 31.41 & 0.0601     \\ 
minority                                 & 1264.92 & 31.75 & 0.0601     \\ 
pessimal                               & 1244.41 & 32.16 & 0.0604     \\
\botrule
\end{tabular}
\footnotetext{Source: The author (2021).}
\end{minipage}
\end{center}
\end{table}

Despite the difference in some of the ranking positions. The positive points obtained in the 70\% cutoff were maintained in the new classification, given the big difference in ranking between artificial and real classifiers. Furthermore, the position of the artificial classifiers was as expected, whereas the Majority classifier is the one with the highest position, followed by the random three, the Minority and ending with the Pessimal.

Another interesting observation was the increase in the maximum rating value and consequently the increase in the difference between Optimal and Random Forest classifiers, which remained as the best real classifier. Furthermore, it is possible to observe a group of classifiers that belong to the same model family. This situation is also consistent with the expected final result, as it is assumed that classifiers from the same family have a similar performance.


The more datasets are filtered for the most discriminating and difficult, the more evident the rating difference between real and artificial classifiers becomes, as can be seen in the ranking with 30\% of the benchmark (refer to Table \ref{tab2}), where the rating difference between Optimal and Random Forest exceeds 100 points.

\begin{table}[h]
\begin{center}
\begin{minipage}{174pt}
\caption{Classifier rating ranking by benchmark subset 30\%.}\label{tab2}%
\begin{tabular}{@{}llll@{}}
\toprule
Classifier & Rating & RD & Volatility\\
\midrule
optimal                                 & 1848.59 & 34.67 & 0.0616     \\ 
RandomForest                         & 1734.63 & 32.64 & 0.0600     \\ 
MLP                & 1697.76 & 31.68 & 0.0601     \\ 
RandomForest(5\_trees) & 1680.40 & 31.59 & 0.0600     \\ 
KNeighbors(3) & 1644.92 & 31.95 & 0.0601     \\ 
KNeighbors(2)              & 1624.71 & 31.73 & 0.0602     \\ 
KNeighbors(5)                                   & 1606.95 & 31.25 & 0.0603     \\ 
KNeighbors(8)               & 1564.72 & 31.33 & 0.0604     \\ 
RandomForest(3\_trees)                            & 1560.64 & 30.96 & 0.0608     \\ 
BernoulliNB               & 1536.59 & 31.50 & 0.0631     \\ 
GaussianNB               & 1531.65 & 31.62 & 0.0637     \\ 
DecisionTree                           & 1529.08 & 31.22 & 0.0620     \\ 
SVM               & 1524.85 & 31.29 & 0.0622     \\ 
majority                           & 1345.71 & 31.77 & 0.0616     \\ 
rand2                                 & 1336.60 & 32.52 & 0.0601     \\ 
minority                                 & 1295.49 & 33.44 & 0.0600     \\ 
rand3                                 & 1292.05 & 32.88 & 0.0601     \\ 
rand1                           & 1254.62 & 34.00 & 0.0602     \\ 
pessimal                               & 1213.69 & 35.12 & 0.0603     \\
\botrule
\end{tabular}
\footnotetext{Source: The author (2021).}
\end{minipage}
\end{center}
\end{table}

However, it is noted that the position of the artificial classifiers has changed. The expected order that was reached by the ranking with a 50\% cutoff has changed and the artificial Minority classifier has dropped from second to the last position and has a rating value very close to that of the Majority classifier. This situation possibly occurs due to the smaller amount of datasets for evaluation and the lower discrimination power in half of the benchmark, as only 18 datasets are used and 9 of them have low discrimination values because they are the most difficult ones. Such condition can be better observed in Table \ref{tab6}.

Note that the set of datasets resulting from the 30\% cut0ff has the lowest mean and highest standard deviation of Discrimination. And simultaneously presents the highest average percentage of Difficulty, but with the highest standard deviation as well. This results in an unbalanced benchmark set that may not be adequate to evaluate models well. As an example, we have the final position of Random Forest with three trees that appeared in the 5th position in the benchmarks with 100\%, 70\% and 50\% of the total datasets, but which performed below the KNN algorithms for the 30\% set.

When analyzing Table \ref{tab6}, it can be seen that the 50\% subset is the one with the most balanced discrimination and difficulty values. Despite not having the greatest discriminatory power in the average, its final value differs only by about 5\% from the highest average, which was reached with the original benchmark. In addition, it is the second highest average difficulty subset at 25.19\%, which also makes it more challenging.


\begin{table}[h]
\begin{center}
\begin{minipage}{174pt}
\caption{Comparison between the Discrimination and Difficulty percentages for each subset.}\label{tab6}%
\begin{tabular}{ccccc}
\toprule
\multicolumn{1}{l}{} & \multicolumn{2}{c}{Discrimination}    & \multicolumn{2}{c}{Difficulty}  \\ 
\midrule
\multicolumn{1}{l}{} & Average  & S. Deviation & Average & S. Deviation         \\ 
\midrule
30\%  & 58.5\%        & 41.91\%            & 33\%         & 33.71\%           \\
50\%  & 62.06\%       & 38.72\%            & 25.19\%      & 28.23\%           \\
70\%  & 65.16\%       & 35.34\%            & 20.44\%      & 25.26\%           \\
100\% & 67.13\%       & 30.78\%            & 15.93\%      & 22.56\%           \\
\botrule
\end{tabular}
\footnotetext{Source: The author (2021).}
\end{minipage}
\end{center}
\end{table}

The empirical analysis of the rating values and the percentages of discrimination and difficulty of each benchmark subset shows that the 50\% cutoff generated the most efficient and balanced subset. This can also be confirmed by evaluating the variance and standard deviation of the RD and Volatility values generated by each rating ranking (refer to Table \ref{tab5}).

For the Glicko system, the lower the RD and Volatility values, the more reliable the ranking and rating values will be. By Table \ref{tab5}, it can be seen that the 30\% subset and the original benchmark feature, respectively, the highest values of variance and standard RD deviation. This means that some classifiers have very high variation ranges, such as the artificial Optimal and Pessimal classifiers. Despite this, the 30\% subset has the smallest volatility variations, revealing an imbalance in the subset.

\begin{table}[h]
\begin{center}
\begin{minipage}{174pt}
\caption{Comparison between RD and Volatility values.}\label{tab5}%
\begin{tabular}{ccccc}
\toprule
\multicolumn{1}{l}{} & \multicolumn{2}{c}{RD}    & \multicolumn{2}{c}{Volatility} \\
\midrule
\multicolumn{1}{l}{} & Variance & S. Deviation & Variance & S. Deviation \\ 
\midrule
30\%                 & 1.19      & 1.49          & 0.0000013 & 0.0011                \\
50\%                 & 0.53      & 0.73          & 0.0000057 & 0.0023                \\
70\%                 & 0.26      & 0.51          & 0.000010  & 0.0032                \\
100\%                & 0.86      & 0.93          & 0.000025  & 0.0050                \\
\botrule
\end{tabular}
\footnotetext{Source: The author (2021).}
\end{minipage}
\end{center}
\end{table}

On the other hand, the subset generated by the 50\% cutoff again shows up as the most consistent, as it has the second smallest RD variation, preceded  only by the 70\% subset. And it also has the second smallest volatility variation, just following the 30\% subset. This allows for inferreing that the 50\% subset generated by decodIRT would be a more efficient and adequate choice than the original OpenML-CC18 benchmark, according to the analysis of the item parameters and the Glicko system.

\subsection{Innate ability evaluation}

In addition to choosing a more efficient benchmark, creating the subsets allows for a deeper assessment of the models' innate ability. It is observed in the newly generated rankings (refer to Tables \ref{tab4}, \ref{tab3} and \ref{tab2}) that Random Forest maintained the lead, ahead of MLP. Like the benchmark subsets, the average difficulty of the benchmarks gradually increases (refer to Table \ref{tab6}). This suggests that RF is probably the most skillful algorithm, as it manages to keep the rating high even in view of the most difficult datasets. Another point that corroborates the previous statement is volatility. In all benchmark scenarios analyzed (100\%, 70\%, 50\% and 30\%), Random Forest is the real classifier with the lowest volatility value, which means that its rating is the most accurate and reliable, so it is less susceptible to possible fluctuation.

The different scenarios also allow for evaluating inverse cases, where the classifiers had a drop in performance. The main example is the SVM, which dropped from 6th to last position among the real classifiers. Although surprising, the Glicko-2 system already pointed out this possibility, since the SVM had the highest volatility value recorded in all the rankings generated, 0.0772. Therefore, this means that SVM was the model with the least reliable rating value.

Although the results are not yet completely conclusive, it is noted that the search for the correct assessment innate ability of the algorithms is valid and that the combination of the use of IRT with the Glicko-2 system can serve as a correct format for this assessment.

\subsection{OpenML-CC18 datasets characterization}

Creating subsets of a benchmark is a very important task, since a smaller benchmark is computationally less costly. However, one should not only consider the computational cost for this task, it is important that the new benchmark generated is also able to maintain characteristics similar to the original.

To test whether the 50\% set generated from the OpenML-CC18 benchmark has similar characterization, the following general dataset characteristics were compiled:

\begin{itemize}
    \item Data types: whether the dataset is composed only of features of numeric or categorical type or whether the dataset is mixed (it has features of both types).
    \item Classification type: whether the dataset is binary or multiclass.
\end{itemize}

\begin{table}[h]
\begin{center}
\begin{minipage}{174pt}
\caption{OpenML-CC18 characterization.}\label{tab7}%
\begin{tabular}{ccccc}
\toprule
& Binary & Multiclass & Total \\ 
\midrule
Categorical              & 6.66\%      & 6.66\%      & 13.33\%  \\
Numeric                & 30\%      & 38.33\%      & 68.33\%  \\
Mixed                   & 11.66\%      & 6.66\%      & 18.33\%  \\
Total                   & 48.33\%      & 51.66\%      & 100\%  \\
\botrule
\end{tabular}
\footnotetext{Source: The author (2021).}
\end{minipage}
\end{center}
\end{table}

\begin{table}[h]
\begin{center}
\begin{minipage}{174pt}
\caption{Characterization of the 50\% subset.}\label{tab8}%
\begin{tabular}{ccccc}
\toprule
& Binary & Multiclass & Total \\ 
\midrule
Categorical              & 6.66\%      & 6.66\%      & 13.33\%  \\
Numeric                & 16.66\%      & 46.66\%      & 63.33\%  \\
Mixed                   & 13.33\%      & 10\%      & 23.33\%  \\
Total                   & 36.66\%      & 63.33\%      & 100\%  \\
\botrule
\end{tabular}
\footnotetext{Source: The author (2021).}
\end{minipage}
\end{center}
\end{table}

When analyzing Tables \ref{tab7} and \ref{tab8}, it can be observed that the data types of the datasets are kept proportional in total between the two sets of benchmarks. Where, for the new benchmark, the total amount of numeric and mixed type datasets varies only by 5\% when compared to the values of the original benchmark.

The most significant change lies in the type of dataset classification, where the original benchmark is more balanced, especially in terms of total values. While the new benchmark generated has a higher total percentage of multiclass datasets, with the difference between the amount of binaries being 26.66\%. However, this wide classification difference only appears when evaluating datasets of numeric data type, since, for categorical datasets, the classification proportion is maintained and for mixed datasets the percentage variation is below 4\% if compared to the original benchmark.

In addition to the general characterization, more specific characteristics of the datasets of each benchmark were also analyzed. To this end, 9 metadata were removed from OpenML for each dataset: number of classes, number of features, percentage of instances with missing values, percentage of categorical features, percentage of numerical features, percentage of binary features, percentage of majority and minority classes and the dimensionality. These data can be accessed in the supplementary material \footnote{Supplementary material: \url{https://osf.io/ytb4d/}}.

Dataset metadata analysis of each set was performed by calculating the mean, median and standard deviation. It was observed whether these values had a significant change when analyzed against the original benchmark and later on the subset generated by the 50\% cutoff. Table \ref{tab9} brings the variation of values between sets in the mean.

\begin{table}[h]
\begin{center}
\begin{minipage}{174pt}
\caption{Variation between the original and the new benchmark in percentage.}\label{tab9}%
\begin{tabular}{ccccc}
\toprule
 & Avg. 100\% & Avg. 50\% \\
\midrule
Nº of Features                      & 151.06  & 136.95   \\
Nº of Classes                       & 6.83 & 5.25  \\
Perc. Instances W. Mis. Val. & 5.64 & 4.97   \\
Perc. Symbolic Features             & 27.24  & 25.77  \\
Perc. Numeric Features              & 72.75   & 74.22   \\
Perc. Binary Features               & 13.10  & 12.23   \\
Perc. Majority Class                & 38.78  & 46.76  \\
Perc. Minority Class                & 16.10  & 19.27    \\
Dimensionality                      & 0.86  & 1.37  \\
\botrule
\end{tabular}
\footnotetext{Source: The author (2021).}
\end{minipage}
\end{center}
\end{table}

Among the analyzed metadata, it is observed that only the Number of Features and the Percentage of the Majority Class undergo considerable variation. Despite the high average number of features, a thorough analysis reveals a standard and median deviation of 335.66 and 25 for the original benchmark and 329.72 and 29 for the new benchmark. This reveals that only a few datasets from both sets have very high numbers of features and that, overall, there is not much variation - see values very close to the median.

The main change is in the percentage of appearance of the majority class, which undergoes a change of about 20\% from the original average value to more in the new benchmark. This shows that the 50\% subset has more datasets with unbalanced classes. Furthermore, a direct analysis of the metadata of each dataset revealed that the highest percentages of the majority class belong to the datasets classified as the most discriminating, i.e. they are the easiest.

It was initially thought that highly unbalanced datasets would be considered the most difficult, but the situation is the opposite. However, this situation is not far from general knowledge in ML. One possible explanation is that the IRT defined that models unable to generalize all classes in an unbalanced situation are bad, while models that actually do are considered good. This would explain the high discrimination values and remain consistent with the natural knowledge of the behavior of models in ML.



\section{Final considerations}

This research work explored the IRT for benchmark evaluation. ML benchmarks are commonly used to explore how far ML algorithms can go when dealing with distinct datasets before selecting the most stable and strong classifier. Although OpenML-CC18 is designed to be a gold standard, it should be used with caution. Out of the 60 datasets evaluated, only 12\% have instances that are considered difficult, while half of the benchmark feature 80\% of the instances as being very discriminatory. This condition can be a great source for analyzing comparisons, but it is not useful for testing the ability of classifiers. The IRT-based benchmark assessment methodology is provided and can be replicated by the decodIRT tool in an automated way. Although classifier skills are highlighted by IRT, there was also a problem with innate skill, whether it is possible to define the boundaries between the ML algorithm (by design) and training (optimization). IRT was exploited in combination with rating systems to establish the ML winner and thereby provide an initial glimpse of a score for what is intended to be called the innate ability of the algorithms. 

In addition, decodIRT was used to explore whether there is a more efficient benchmark subset than the original one and whether more information could be obtained about model ability and data complexity from the IRT estimators. After exploring different subsets, the subset consisting of 50\% of the total datasets selected on the percentages of discrimination and difficulty was chosen. This subset features dataset properties which are very close to the original ones, but it also proved to be more suitable for evaluating and separating the strength of the models. Furthermore, the creation of these benchmark subsets allowed for exploring the innate ability of the models. The final result pointed out that Random Forest is the classifier that has the greatest innate ability, and chosing it is preferable in relation to the other evaluated models while facing distinct datasets complexities. Therefore, it was shown that the IRT can also be used for filtering and creating more efficient sets of datasets from benchmarks. So, in this ``dispute'' between data vs classifiers, the final result was a technical draw to decide which is the most important, though data complexity always comes first.

Machine learning is a challenge loop of improving data and models to obtain better results. This research work introduces a methodology and tool based on IRT to guide more in-depth exploration on this challenge. The assumption that a given dataset is more difficult than another must be made with caution, since it is directly linked to the inner ability of models (classifiers). The number of classifiers used in a benchmark evaluation should take into account distinct families of classifers. The evaluation of classifiers is also a key point that can be better explored by the utilization of rating systems as Glicko-2.


\bmhead{Acknowledgments}

The authors hereby would like to thank the Federal University of Pará (UFPA), the Graduate Program in Computer Science (PPGCC) and the Vale Technological Institute (ITV) for supporting development and research. This research work had financial support from the Brazilian agency CNPq (Conselho Nacional de Desenvolvimento Científico e Tecnológico) and was supported by Vale (Genomics Biodiversity project, Grant No. RBRS000603.85) to Ronnie Alves. The funders had no role in the study design, data collection and interpretation, or the decision to submit the work for publication.

\bibliography{sn-bibliography}


\end{document}